\def\cvprPaperID{9733}
\def\confName{CVPR}
\def\confYear{2023}

\def\paperTitle{Learning to Render Novel Views from Wide-Baseline Stereo Pairs}

\def\authorBlock{
    Yilun Du \qquad
    Cameron Smith \qquad
    Ayush Tewari\footnotemark[2] \qquad
    Vincent Sitzmann\footnotemark[2] \\ 
    MIT CSAIL \\
    {\tt\small \{yilundu, camsmith, ayusht, sitzmann\}@mit.edu} 
}

\newif\ifreview 
\newif\ifarxiv \newcommand{\arxiv}{\arxivtrue}
\newif\ifcamera \newcommand{\cameraready}{\cameratrue}
\newif\ifrebuttal 

\cameraready %
\arxiv

\pdfoutput=1
\documentclass[10pt,twocolumn,letterpaper]{article}
\ifreview \usepackage[review]{cvpr} \fi
\ifarxiv \usepackage[pagenumbers]{cvpr} \fi
\ifrebuttal \usepackage[rebuttal]{cvpr} \fi
\ifcamera \usepackage{cvpr} \fi

\usepackage[accsupp]{axessibility}
\usepackage{graphicx}
\usepackage{amsmath}
\usepackage{amssymb}
\usepackage{booktabs}
\usepackage[dvipsnames]{xcolor}

\usepackage{times}
\usepackage{microtype}
\usepackage{epsfig}
\usepackage{caption}
\usepackage{float}
\usepackage{placeins}
\usepackage{color, colortbl}
\usepackage{stfloats}
\usepackage{enumitem}
\usepackage{tabularx}
\usepackage{xstring}
\usepackage{multirow}
\usepackage{xspace}
\usepackage{url}
\usepackage{subcaption}
\usepackage[hang,flushmargin]{footmisc}

\ifcamera \usepackage[accsupp]{axessibility} \fi

\ifarxiv  \fi

\newcommand{\ours}[1]{\textcolor{cyan}{OURS}\xspace}

\newcommand{\R}[1]{{%
    \textbf{%
        \ifstrequal{#1}{1}{\textcolor{red}{R#1}}{%
        \ifstrequal{#1}{2}{\textcolor{blue}{R#1}}{%
        \ifstrequal{#1}{3}{\textcolor{magenta}{R#1}}{%
        \ifstrequal{#1}{4}{\textcolor{teal}{R#1}}{%
                           \textcolor{cyan}{R#1}%
        }}}}%
    }%
}}

\newcommand{\myparagraph}[1]{\vspace{-8pt}\paragraph{#1}}

\newcommand{\sect}[1]{Section~\ref{#1}}

\newcommand{\fig}[1]{Figure~\ref{#1}}
\newcommand{\tbl}[1]{Table~\ref{#1}}

\definecolor{tabbestcolor}{rgb}{0.785, 0.851, 0.969}
\def \best {\cellcolor{tabbestcolor!85}}
\def \sbest {\cellcolor{tabbestcolor!30}}

\DeclareMathOperator*{\argmin}{arg\,min}  %

\usepackage{xr-hyper}

\makeatletter
\newcommand*{\addFileDependency}[1]{
  \typeout{(#1)}
  \@addtofilelist{#1}
  \IfFileExists{#1}{}{\typeout{No file #1.}}
}

\makeatother

\usepackage[pagebackref,breaklinks,colorlinks]{hyperref}
\usepackage[capitalize]{cleveref}
\crefname{section}{Sec.}{Secs.}
\crefname{table}{Table}{Tables}
\crefname{figure}{Fig.}{Figs.}

\begin{document}
\title{\paperTitle}
\author{\authorBlock}

\twocolumn[{%
\renewcommand\twocolumn[1][]{#1}%
\maketitle
\begin{center}
    \vspace{-3mm}
    \centering
    \captionsetup{type=figure}
    \includegraphics[width=\linewidth]{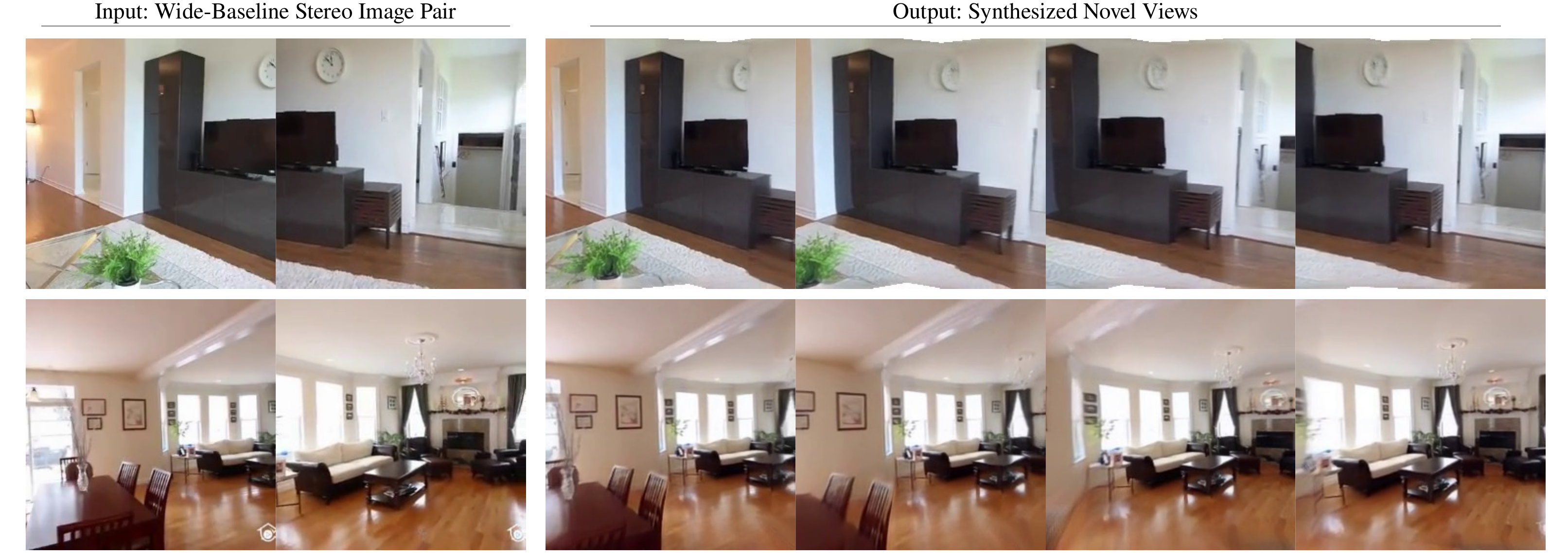}
    \vspace{-7mm}
    \captionof{figure}{
    \textbf{Novel view synthesis from a single wide-baseline stereo image pair.} In a single forward pass, our method maps a wide-baseline stereo image pair to features that enable fast rendering of novel views, trained using only posed multi-view images of static scenes without ground-truth or proxy geometry. We outperform all prior art on novel view synthesis from sparse observations, taking a significant step towards matching the quality of overfitting on single scenes in this challenging setting.
    }
    \label{fig:teaser}
\end{center}
}]

\let\thefootnote\relax\footnotetext{$\dagger$ Equal Advising \\ Project website: \url{https://yilundu.github.io/wide_baseline/}}
\begin{abstract}
We introduce a method for novel view synthesis given only a single wide-baseline stereo image pair.
In this challenging regime, 3D scene points are regularly observed only once, requiring prior-based reconstruction of scene geometry and appearance.
We find that existing approaches to novel view synthesis from sparse observations fail due to recovering incorrect 3D geometry and due to the high cost of differentiable rendering that precludes their scaling to large-scale training.
We take a step towards resolving these shortcomings by formulating a multi-view transformer encoder, proposing an efficient, image-space epipolar line sampling scheme to assemble image features for a target ray, and a lightweight cross-attention-based renderer.
Our contributions enable training of our method on a large-scale real-world dataset of indoor and outdoor scenes.
We demonstrate that our method learns powerful multi-view geometry priors while reducing the rendering time. 
We conduct extensive comparisons on held-out test scenes across two real-world datasets, significantly outperforming prior work on novel view synthesis from sparse image observations and achieving multi-view-consistent novel view synthesis.
\end{abstract}

\section{Introduction}
\label{sec:intro}

The goal of novel view synthesis is to render images of a scene from unseen camera viewpoints given a set of image observations. 
In recent years, the emergence of differentiable rendering~\cite{mildenhall2020nerf,lombardi2019neural,sitzmann2019deepvoxels,sitzmann2019scene,tewari2022advances} has led to a leap in quality and applicability of these approaches, enabling near photorealistic results for most real-world 3D scenes.
However, methods that approach photorealism require hundreds or even thousands of images carefully exploring every part of the scene,  where special care must be taken by the user to densely image all 3D points in the scene from multiple angles. 

In contrast, we are interested in the regime of novel view synthesis from a sparse set of context views. 
Specifically, this paper explores whether it is possible to sythesize novel view images using an extremely sparse set of observations. 
In the most challenging case, this problem reduces to using input images such that every 3D point in the scene is only observed from a \emph{single} camera perspective.
Towards this goal, we propose a system that uses only a single wide-baseline stereo image pair of the scene as input. 
This stereo image pair regularly has little overlap, such that many 3D points are indeed only observed in one of the images, see Fig.~\ref{fig:teaser}. 
Image observations themselves are thus insufficient information to compute 3D geometry and appearance via multi-view stereo, and we must instead \emph{learn} prior-based 3D reconstruction. Nevertheless, reasoning about multi-view consistency is critical, as prior-based reconstructions must agree across images to ensure multi-view-consistent reconstruction.

This is a novel problem setting: 
While some existing methods demonstrate novel view synthesis from very sparse observations~\cite{pixelnerf,trevithick2021grf,sitzmann2019scene}, they are limited to object-level scenes. 
In contrast, we are interested in large real-world scenes that are composed of multiple objects with complex geometry and occlusions. 
Previous approaches for novel view synthesis of scenes focus on small baseline renderings using $3-10$ images as input~\cite{wang2021ibrnet,suhail2022generalizable,SRF,chen2021mvsnerf,johari2022geonerf,liu2022neural, pixelnerf}. 
In this setting, most 3D points in the scene are observed in multiple input images, and multi-view feature correspondences can be used to regress 3D geometry and appearance.
Thus, these methods in practice learn to amortize multi-view stereo.
In our setting, we use a wide-baseline stereo image pair as input, where it is not sufficient to rely on multi-view feature correspondences due to many points only being observed in a single view.
We show that in this challenging setting, existing approaches do not faithfully recover the 3D geometry of the scene. 
In addition, most existing methods rely on costly volume rendering for novel view synthesis, where the number of samples per ray required for high-quality rendering makes it difficult to train on complex real-world scenes.

In this paper, we propose a new method that addresses these limitations, and provides the first solution for high-quality novel 
 view synthesis of a scene from a wide-baseline stereo image pair.  
To better reason about the 3D scene, we introduce a multi-view vision transformer that computes pixel-aligned features for each input image. 
In contrast to a monocular image encoder commonly used in previous approaches~\cite{pixelnerf,wang2021ibrnet,trevithick2021grf}, the multi-view transformer uses the camera pose information as input to better reason about the scene geometry. 
We reduce the memory and computational costs for computing image features by combining this vision transformer at lower resolutions with a CNN at higher resolutions.
A multi-view feature matching step further refines the geometry encoded in these feature maps for any 3D point that can be observed in both images. 

We also introduce an efficient differentiable renderer that enables large-scale training. 
Existing approaches that use volume rendering sample points along camera rays in 3D and project these points onto the image planes to compute the corresponding features using bilinear interpolation. 
Since perspective projection is a non-linear operation, uniformly sampled 3D points are not uniformly distributed in 2D, leading to some pixels in the feature maps being sampled multiple times, and other pixels not being sampled at all. 
Thus, this sampling strategy does not use the information in the pixel-aligned feature maps optimally. 
We instead take an image-centric sampling approach where we first compute the epipolar lines of a target pixel in the input images, and sample points uniformly on these lines in 2D. 
This exploits the fact that the number of pixels along the epipolar lines is the maximum effective number of samples. 
In addition, we use lightweight cross-attention layers that directly aggregate the sampled features and compute the pixel color.
In contrast to volume rendering where we need to sample very close to a surface in order to render its color, thus requiring a large number of samples, our learned renderer does not share this limitation and can compute the pixel color even with sparse samples. 
Our lightweight rendering and feature backbone components enable us to train on large-scale real-world datasets. 
We demonstrate through extensive experiments on two datasets that our method achieves state-of-the-art results, significantly outperforming existing approaches for novel view synthesis from sparse inputs.

\vspace{-3pt}
\section{Related Work}
\label{sec:related}

\paragraph{Image-based rendering.} 
Image-based rendering (IBR) methods generate images from novel camera viewpoints by blending information from a set of input images. 
We provide a brief overview of some methods. 
Please refer to the review by Shum and Kang~\cite{shum2000review} for details. 
Some IBR approaches directly model the plenoptic function without using information about the scene geometry~\cite{levoy1996light, ng2005light}. 
Other approaches use a proxy scene geometry computed using multi-view stereo to guide the blending of information from the input images~\cite{debevec1996modeling,buehler2001unstructured,heigl1999plenoptic,liu2009content}. 
While rendering without computing an explicit 3D geometry leads to higher-quality results, it requires a large number of input images.
In contrast, methods that rely on 3D geometry can work with sparse image inputs. 
However, multi-view stereo from a sparse set of input views often leads to inaccurate geometry, especially for scenes with complex geometry, limiting the quality of rendered images. 
Methods have been proposed for higher-quality geometry computation~\cite{chaurasia2013depth,hedman2016scalable}, optical flow-based refinement~\cite{eisemann2008floating,casas20154d,du2018montage4d}, and improved blending~\cite{penner2017soft,hedman2018deep,Riegler2020FVS}. 
In contrast to these image-based rendering methods, we rely on priors learned from data that enable novel-view synthesis from just a wide-baseline stereo image. 
We do not create any explicit proxy geometry of the scene and are thus unaffected by inaccurate multi-view stereo.

\myparagraph{Single-Scene Volumetric Approaches.}
Recent progress in neural rendering~\cite{tewari2022advances} and neural fields ~\cite{sitzmann2019siren, mildenhall2020nerf, xie2021neural} has led to a drastic jump in the quality of novel-view synthesis from several input images of a scene. 
Here, a 3D scene representation is optimized via differentiable rendering to fit a set of image observations. 
Early approaches leveraged voxel grids and learned renderers~\cite{nguyen2018rendernet,sitzmann2019deepvoxels,lombardi2019neural}.  
More recent approaches rely on neural fields~\cite{xie2021neural,mildenhall2020nerf,barron2021mipnerf,martin2021nerf} to parameterize the 3D scene and volumetric rendering~\cite{tagliasacchi2022volume,mildenhall2020nerf,lombardi2019neural} for image synthesis. 
This leads to photorealistic view synthesis but requires hundreds of input images that densely sample the 3D scene. 
Hand-crafted and learned priors may reduce the number of required images to the order of three to ten~\cite{niemeyer2022regnerf}, but 3D points still need to be observed from at least two perspectives.
A major challenge of these approaches is the cost of accurate differentiable rendering, regularly requiring hundreds  of samples per ray. Recent approaches have achieved impressive speed-ups in 3D reconstruction leveraging high-performance data structures and sparsity~\cite{mueller2022instant,fridovich2022plenoxels, chen2022tensorf,liu2020neural}. 
While promising, reconstruction can still take a few minutes per scene, and sparse data structures such as octrees and hash tables cannot easily be used  with learned priors.

Our approach tackles a different setting than these methods, using only a single wide-baseline stereo image as input, where 3D points are regularly only observed in a \emph{single} view. 
Our approach does not require any per-scene optimization at test time. 
Instead, it reconstructs the scene in a single forward pass.
Note that while our method does not achieve the quality of per-scene optimization methods that use hundreds of input images, it demonstrates a significant step up in novel view synthesis from very sparse image observations. 

\myparagraph{Prior-based 3D Reconstruction and View Synthesis.}
Instead of overfitting to a single scene, differentiable rendering can also be used to supervise prior-based inference methods.
Some methods generalize image-based rendering techniques by computing feature maps on top of a proxy geometry~\cite{Riegler2020FVS,aliev2020neural,kopanas2021point,wiles2020synsin}.
Volume rendering using multi-plane images has been used for small baseline novel view synthesis~\cite{realestate10k,srinivasan2019pushing,tucker2020single,zhang2022video}. 
Early neural fields-based approaches~\cite{sitzmann2019scene, niemeyer2020differentiable} were conditioned on a single global latent code and rendered via sphere tracing. 
In contrast to a global latent code, several approaches use a feature backbone to compute pixel-aligned features that can be transformed using MLPs~\cite{pixelnerf,lin2022vision,trevithick2021grf} or transformers layers~\cite{wang2021ibrnet,reizenstein2021common} to a radiance field.
Ideas from multi-view stereo such as the construction of plane-swept cost volumes~\cite{chen2021mvsnerf,liu2022neural,johari2022geonerf}, or multi-view feature matching~\cite{SRF} have been used for higher-quality results.

Alternatively to these radiance field-based approaches, some methods use a light field rendering formulation where an oriented camera ray can directly be transformed to the pixel color as a function of the features computed from the input images~\cite{sitzmann2021light,suhail2022light}. 
Scene Representation Transformers~\cite{sajjadi2021scene} use transformers with global attention to compute a set-latent representation that can be decoded to pixel colors when queried with a target camera ray. 
However, global attention layers on high-resolution input images are very compute and memory intensive. 
Developed concurrently with our work, Suhail~\etal~\cite{suhail2022generalizable} proposed to use a transformer to only compute features for image patches along the epipolar rays of the pixel being rendered.
This is still very expensive due to global attention layer computations over multiple image patches for every rendered pixel. 
In addition, this method ignores the context information of the scene, since all computation is performed only for patches that lie on the epipolar lines.

All existing prior-based reconstruction methods either only support object-level scenes or very small baseline renderings,  or rely on multiple image observations where most 3D points are observed in multiple input images.
This is different from our setting where we only use a wide-baseline stereo image pair of scenes as input.

\begin{figure*}
    \centering
    \includegraphics[width=\linewidth]{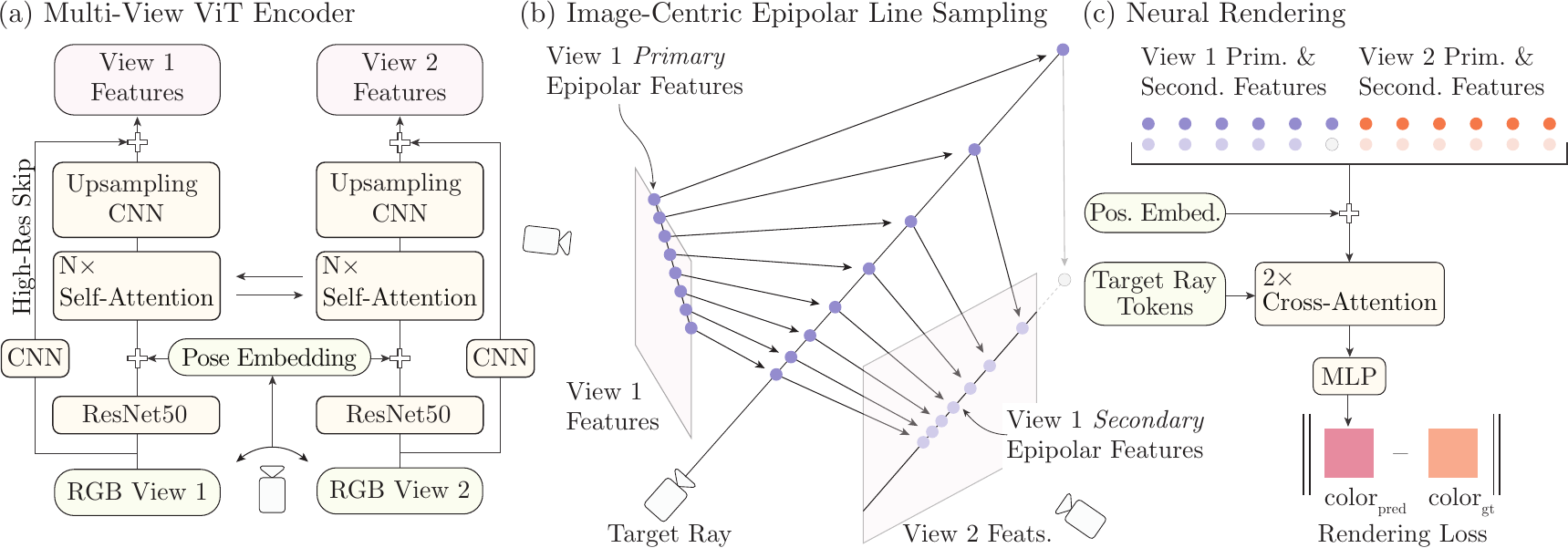}
    \caption{\textbf{Method Overview.} 
    (a) Given context images from different viewpoints, a multi-view encoder extracts pixel-aligned features, leveraging attention across the images and their corresponding camera pose embeddings. 
    (b) Given a target ray, in each context view, we sample \emph{primary} features along the epipolar line equidistant in pixel space. We then project the corresponding 3D points onto the other views and sample corresponding \emph{secondary} epipolar line features, where out-of-bounds features are set to zero. 
    (c) We render the target ray by performing cross-attention over the set of all primary and secondary epipolar line features from all views.
    }
    \label{fig:overview}
    \vspace{-10pt}
\end{figure*}

\section{Method}
\label{sec:method}

Our goal is to render novel views of a 3D scene given a wide-baseline stereo image pair $\mathcal{I}_1$ and $\mathcal{I}_2$.
We assume known camera intrinsic $\mathbf{K}_i \in \mathbb{R}^{3 \times 3}$ and extrinsic $\mathbf{E}_i \in \mathbb{R}^{4 \times 3}$ expressed relative to context camera 1.
We use a multi-view encoder to compute pixel-aligned features, and a cross-attention-based renderer to transform the features into novel view renderings, see \fig{fig:overview} for an overview. 

\subsection{Multiview Feature Encoding}
An essential part of novel view synthesis given context images is an accurate reconstruction of scene geometry. 
Our method implicitly reconstructs 3D geometry and appearance of the scene in the form of pixel-aligned feature maps for each stereo image.
In prior work, pixel-aligned features are obtained by separately encoding each image via a vision transformer or CNN~\cite{pixelnerf,lin2022vision}.
However, in our early experiments, we found this led to artifacts in renderings observing boundary regions between context images. 
We hypothesize that separate encoding of images leads to inconsistent geometry reconstruction across context images.
We thus introduce our \emph{multi-view encoder}, which obtains pixel-aligned features by \emph{jointly} processing the images and the relative pose between them.
Encoding the pose information has also been shown to act as 
an effective inductive bias for 3D tasks~\cite{yifan2022input}.

We now describe this architecture in detail, which extends the dense vision transformer proposed by Ranftl et al.~\cite{ranftl2021vision}. Please see \fig{fig:overview} for an overview.
From each stereo image, we first independently extract convolutional features via a ResNet50 CNN.
We then flatten both images, obtaining $2 \times 16 \times 16$ features in total.
To each feature, we add (1) a learned per-pixel positional embedding encoding its pixel coordinate and (2) a camera pose embedding, obtained via a linear transform of the relative camera pose between context images 1 and 2.
These tokens are processed by a vision transformer, which critically performs self-attention across \emph{all} tokens across \emph{both} images.
In-between self-attention layers, per-image features are re-assembled into a spatial grid, up-sampled, and processed by a fusion CNN~\cite{ranftl2021vision} to yield per-image spatial feature map. 
Directly using these spatial feature maps for novel view synthesis leads to blurry reconstructions, due to the loss of high-frequency texture information. 
We thus concatenate these features with high-resolution image features obtained from a shallow CNN.

\subsection{Epipolar Line Sampling and Feature Matching}
We aim to render an image of the scene encoded in the two pixel-aligned feature maps from a novel camera viewpoint. 
A common way to achieve this is volume rendering, where we cast a camera ray, compute density and color values at many depths along the ray, and integrate them to compute the color of the pixel. 
Sampling locations are determined in 3D. 
Coarse samples are either uniformly spaced in euclidean space or spaced with uniform disparity, and fine samples are distributed closer to the surface as computed by the coarse samples~\cite{mildenhall2020nerf,barron2021mipnerf,donerf}.
However, in our regime of generalizable novel view synthesis with pixel-aligned feature maps, this sampling scheme is suboptimal. 
In this case, sampling along the ray should be determined by the resolution of the context images: the number of pixels along the epipolar line is the maximum effective number of samples available for any method. 
More samples would not provide any extra information.
We propose a sampling strategy to exploit this and demonstrate its effectiveness in an ablation study.

Consider a pixel coordinate $\textbf{u}_t = (u,v)$ in the target image $\mathcal{I}_t$, with assumed known intrinsic $\mathbf{K}_t$ and extrinsic $\textbf{T}_t= \left[
\begin{smallmatrix}
\mathbf{R}_t & \textbf{t}_t \\
\textbf{0} & 1
\end{smallmatrix}\right]$
camera parameters relative to the context camera $\mathcal{I}_1$.
Its epipolar lines $\mathbf{l}_{\{1,2\}}$, in context cameras 1 and 2 are given as:
\begin{align}
\small
\mathbf{l}_i = \mathbf{F}_i \left[u,v,1\right]^T = \mathbf{K}_i^{-T}([\mathbf{t}_t]_\times\mathbf{R}_t)\mathbf{K}_t^{-1} \left[u,v,1\right]^T 
\end{align}
via the fundamental matrix $\mathbf{F}_i$. 
We now uniformly sample $N$ pixel coordinates along the line segment of the epipolar line within the image boundaries. 
To enable the renderer to reason about whether to use a certain pixel-aligned feature or not, a critical piece of information is the depth in the context coordinate frame at which we are sampling this feature.
This depth value can be computed via triangulation, using a closed-form expression. 
Please refer to the supplemental document for details. 
We now obtain $N$ tuples $\{(d, \mathbf{f})_k\}_{k=1}^N$ of depth $d$ and image feature $\mathbf{f}$ per context image for a total of $2N$ samples which we call \emph{primary} samples.

We further propose a feature matching module to refine the geometry encoded in the primary epipolar line samples via correspondence matching.
Consider a primary epipolar line sample obtained from context image $i$, a tuple $(d, \mathbf{f})$ corresponding to a pixel coordinate $\mathbf{u}_t$. 
We propose to augment this sample by a corresponding feature in the other context image. Specifically, we first solve for the corresponding 3D point, and then project this 3D point onto the \emph{other} context image to retrieve a corresponding feature $\hat{\mathbf{f}}$, which we refer to as a \emph{secondary} feature.
The secondary features are set to zero if the projected point is out of the image bounds.
Intuitively, primary and secondary features \emph{together} allow a final stage of geometry refinement for 3D points that are observed in both images: if the features agree, this sample likely encodes a surface.
If the projected point on the other image lies outside the image boundary, we simply set the secondary features to zeros. 
We obtain the input to the renderer as the final set of features by concatenating each primary epipolar line feature with its corresponding secondary feature in the other context view, yielding a set $\{(d, \mathbf{f}, \hat{\mathbf{f}})_k)_{k=1}^{2N}$.
In practice, we sample $N = 64$ points on the epipolar lines for both images, leading to a total of $2N = 128$ tuples. 

\subsection{Differentiable Rendering via Cross-Attention}
To render the target ray, it remains to map the set of epipolar line samples $\{(d, \mathbf{f}, \hat{\mathbf{f}})_k)_{k=1}^{2N}$ to a color value.
As this operation has to be executed once per ray, a key consideration in the design of this function is computational cost.
We propose to perform rendering via a lightweight cross-attention decoder.

For each point on the epipolar line, we embed the target ray origin $\mathbf{o}_t$, target ray direction $\mathbf{r}_t$,
depth with respect to the target ray origin $d_t$, and context camera ray direction $\mathbf{r}_c$ for the epipolar point into a ray query token $\mathbf{q}$ via a shallow MLP as $\Phi([\mathbf{o}_t, \mathbf{r}_t, \mathbf{r}_c, d_t])$.
The $2N$ ray feature values are independently transformed into key and value tokens using a 2-layer MLP. 
Our renderer now performs two rounds of cross-attention over this set of features to obtain a final feature embedding, which is then decoded into color via a small MLP.

The expectation of the Softmax distribution over the sampled features gives a rough idea of the scene depth as
    $e = \sum_k d_k \alpha_k$,
where $d_k$ denotes the depth of the $k$-th epipolar ray sample along the target ray and $\alpha_k$ is the corresponding Softmax weight as computed by the cross-attention operator.
Note that $e$ is not the actual depth but a measure of which epipolar samples the renderer uses to compute the pixel color. 
Unlike volume rendering, where we need to sample very close to a surface to render its color, our light field-based renderer can reason about the surface without exactly sampling on it. 
The learned cross-attention layers can use the target camera ray information, along with a sparse set of epipolar samples, to compute the pixel color. Thus, our method does not require explicit computation of accurate scene depth for rendering.

\subsection{Training and Losses}
We now have a rendered image from a novel camera viewpoint. 
Our loss function consists of two terms: 
\begin{equation}
    \mathcal{L} = \mathcal{L}_\text{img} + \lambda_\text{reg} \mathcal{L}_\text{reg} \,.
\end{equation}
The first term evaluates the difference between the rendered image from a novel camera viewpoint, $R$ and the ground truth, $G$ as:
\begin{equation}
    \mathcal{L}_\text{img} = || R - G ||_1 + \lambda_\text{LPIPS} \mathcal{L}_\text{LPIPS} (R, G) \,,
\end{equation}
where $\mathcal{L}_\text{LPIPS}$ is the LPIPS perceptual loss~\cite{Zhang2018Unreasonable}. 
In practice, we render square patches with a length of $32$ pixels and evaluate these image losses at the patch level. 

We also use a regularization loss on the cross-attention weights of the renderer for better multi-view consistency: 
\begin{equation}
    \mathcal{L}_\text{reg} = \sum_{(u,v)} \sum_{(u'v') \in \mathcal{N}(u,v)} ((e(u,v) - e(u',v'))^2 \,.
\end{equation}
Here, $e(u,v)$ denotes the expected value of the depth of the epipolar samples at pixel $(u,v)$, and $\mathcal{N}()$ defines the neighborhood around a pixel.

For better generalization, we further perform several geometrically-consistent data augmentations during the training procedure. 
We center crop and scale the input and target images, which leads to transformation in the intrinsics of the camera. 
We also flip the images which leads to transformation of the extrinsics.

\section{Experiments}
\label{sec:exp}

We quantitatively and qualitatively show that our approach can effectively render novel views from wide-baseline stereo pairs.
We describe our underlying experimental setup in \sect{sect:experimental}. 
Next, we evaluate our approach on challenging indoor scenes with substantial occlusions in \sect{sect:indoor}. 
We further evaluate on outdoor scenes in \sect{sect:outdoor}. 
We analyze and ablate the underlying components in \sect{sect:analysis}. 
Finally, we illustrate how our approach can render novel views of unposed images of scenes captured in the wild in \sect{sect:in_the_wild}.

\subsection{Experimental Setup}
\label{sect:experimental}

\paragraph{Datasets.} We train and evaluate our approach on RealEstate10k~\cite{realestate10k}, a large dataset of indoor and outdoor scenes, and ACID~\cite{infinite_nature_2020}, a large dataset of outdoor scenes. 
We use 67477 scenes for training and 7289 scenes for testing for RealEstate10k, and 11075 scenes for training and 1972 scenes for testing for ACID, following default splits. 
We train our method on images at $256 \times 256$ resolution and evaluate methods on their ability to reconstruct intermediate views in test scenes (details in the supplement).

\myparagraph{Baselines.} We compare  to several existing approaches for novel view synthesis from sparse image observations. 
We compare to pixelNeRF~\cite{pixelnerf} and IBRNet~\cite{wang2021ibrnet} that use pixel-aligned features, which are decoded into 3D volumes  rendered using volumetric rendering. 
We also compare to Generalizable Patch-based Rendering (GPNR)~\cite{suhail2022generalizable}, which uses a vision transformer-based backbone to compute epipolar features, and a light field-based renderer to compute pixel colors. 
These baselines cover a wide range of design choices used in existing methods, such as pixel-aligned feature maps computed using CNNs~\cite{pixelnerf,wang2021ibrnet} and transformers~\cite{suhail2022generalizable}, volumetric rendering by decoding features using MLPs~\cite{pixelnerf} and transformers~\cite{wang2021ibrnet}, and light field-based rendering~\cite{suhail2022generalizable}.
We use publicly available codebases for all baselines and train them on the same datasets we use for fair evaluations. 
Please refer to the supplemental for comparisons to more baselines. 

\myparagraph{Evaluation Metrics.} We use LPIPS~\cite{Zhang2018Unreasonable}, PSNR, SSIM~\cite{Wang2004Image}, and MSE metrics to compare the image quality of rendered images with the ground truth.

\subsection{Indoor Scene Neural Rendering}
\label{sect:indoor}

We first evaluate the ability of our approach and baselines to render novel views in complex indoor environments with substantial occlusions between objects.

\begin{figure*}[t]
    \centering
    \includegraphics[width=\linewidth]{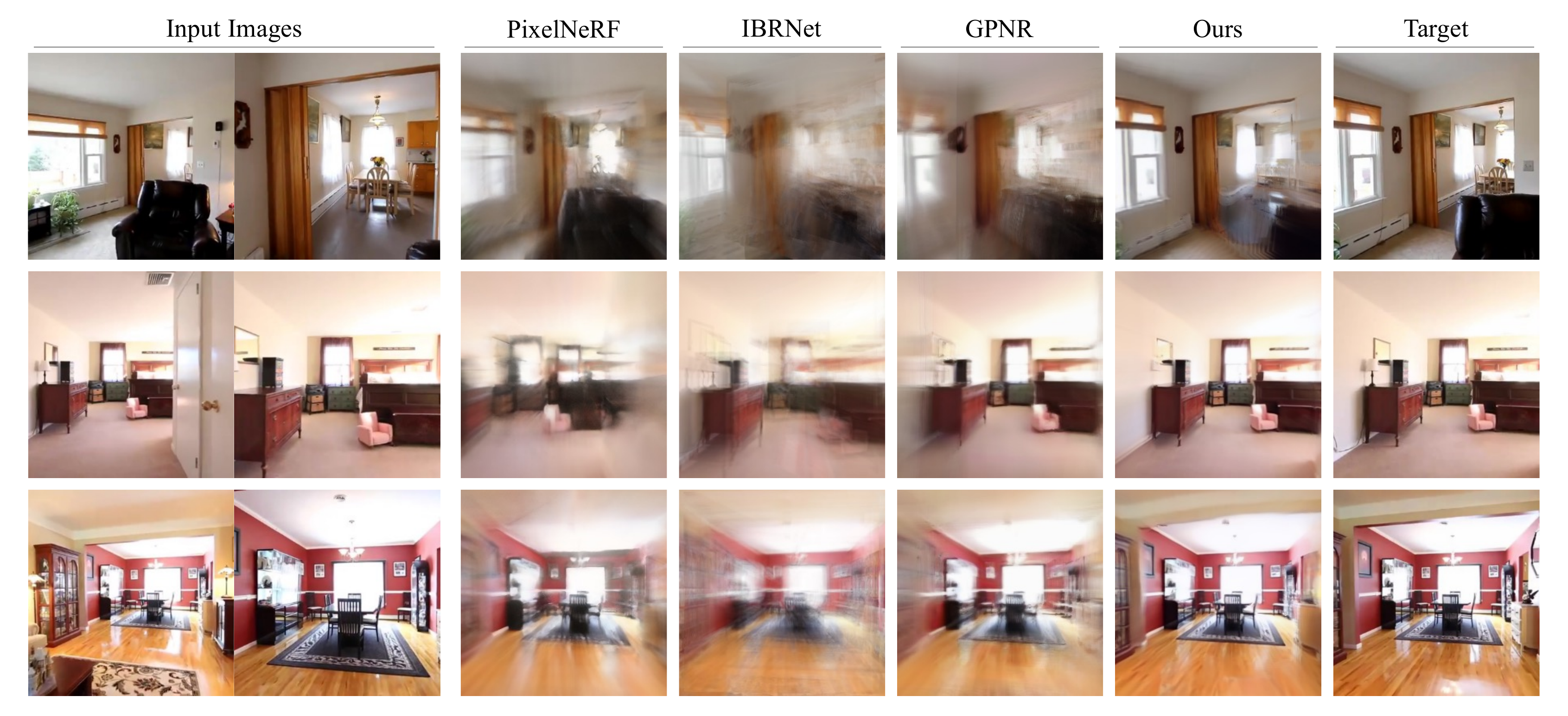}
    \vspace{-10pt}
    \caption{\textbf{Comparative Rendering Results on RealEstate10k.} Our approach can render novel views of indoor scenes with substantial occlusions with high fidelity using a wide-baseline input image pair, outperforming all baselines. 
    Note that many points of the 3D scene are only observed in a single image in such inputs.
    Our method can correctly reason about the 3D structures from such sparse views. 
    }
    \label{fig:realestate}
\end{figure*}
\begin{figure}[t]
    \centering
    \includegraphics[width=\linewidth]{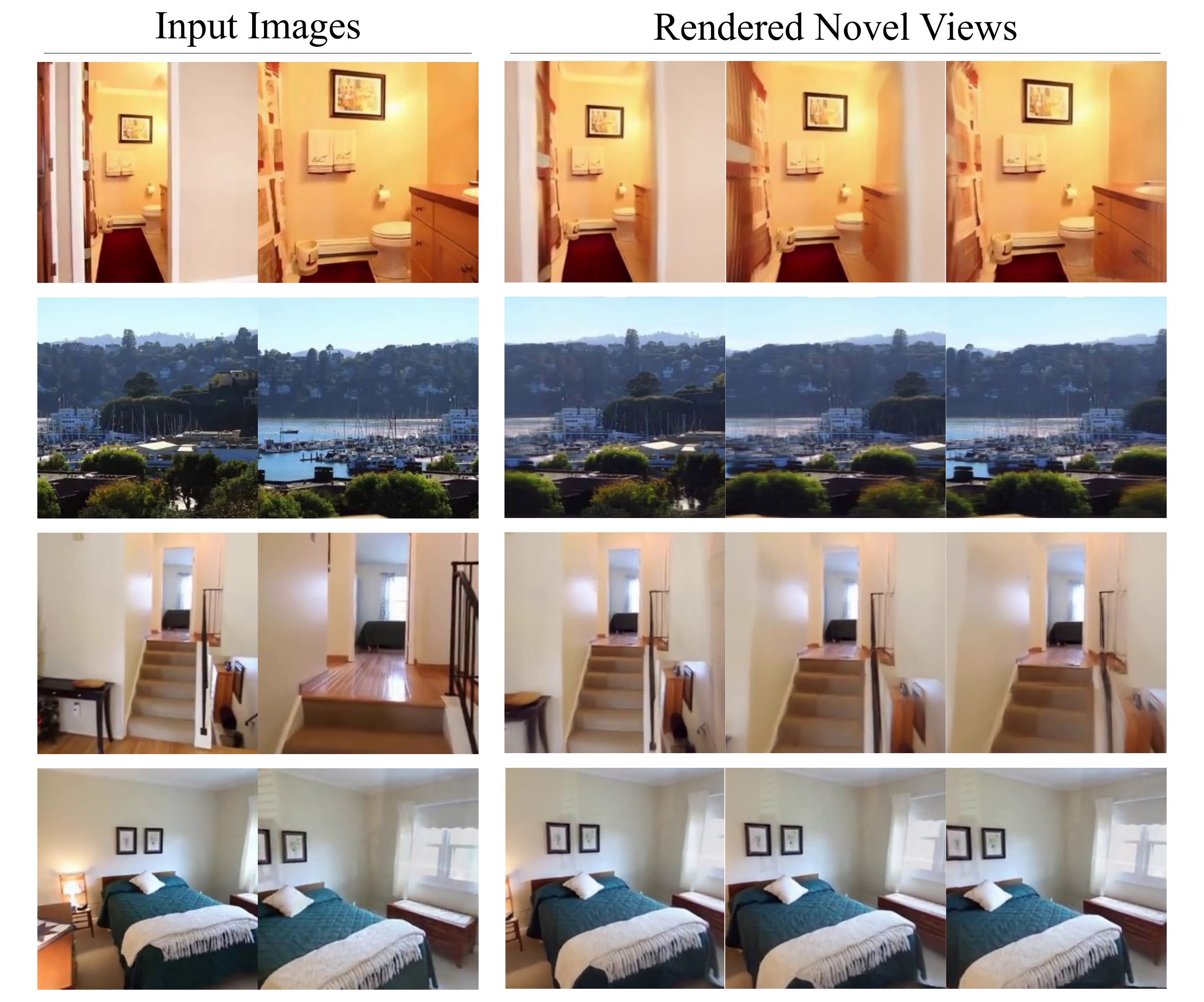}
    \caption{\textbf{ Novel view renderings of our approach given a large baseline stereo pair.} Our approach can synthesize intermediate views that are substantially different from input images, even with very limited overlap between images.
    }
    \label{fig:realestate_more}
    \vspace{-10pt}
\end{figure}
\begin{figure*}[t]
    \centering
    \includegraphics[width=0.9\linewidth]{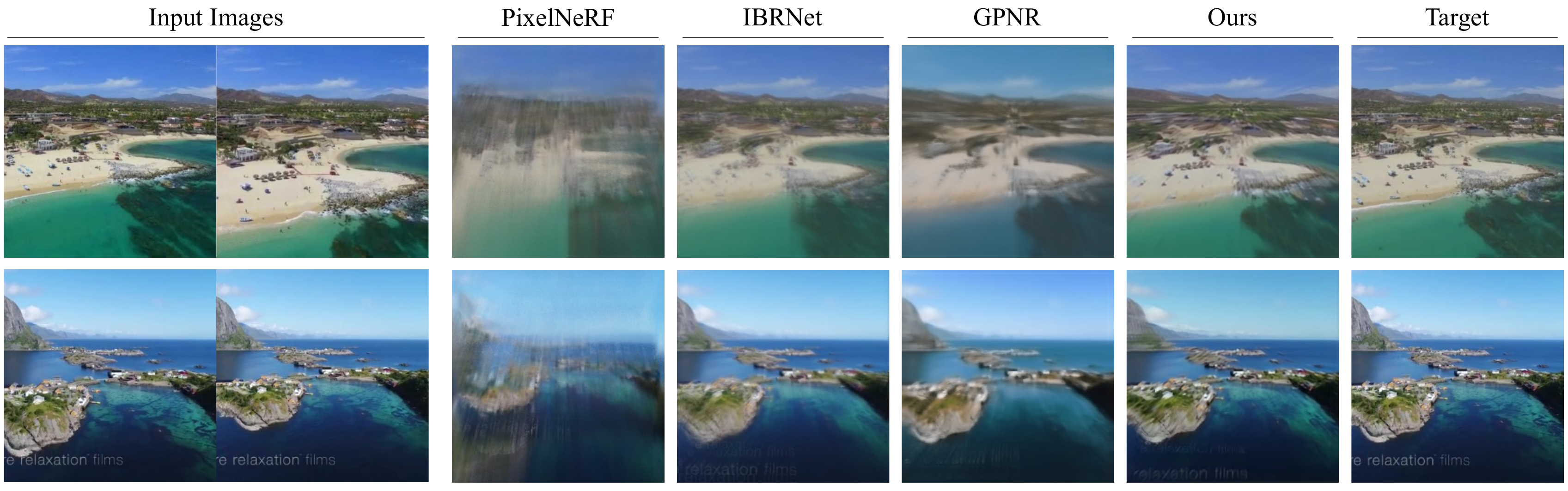}
    \vspace{-5pt}
    \caption{\textbf{Comparative Results on ACID.} Our approach is able to render novels views with higher quality than all baselines. }
    \label{fig:acid}
    \vspace{-5pt}
\end{figure*}

\myparagraph{Qualitative Results.} In \fig{fig:realestate}, we provide qualitative results of novel view renderings of our approach, compared to each of our baselines. 
We provide additional novel view results of our method in \fig{fig:realestate_more}. 
Compared to the baselines, our approach reconstructs the 3D structure of the scene better, and also captures more high-frequency details.  

\myparagraph{Quantitative Results.} We quantitatively evaluate our approach and baselines in \tbl{tab:realestate}. We find that our approach substantially outperforms each compared baseline in terms of all of our metrics.

\begin{table}
\centering
\scalebox{0.9}{%

\begin{tabular}{lccccc}
\toprule
Method & LPIPS $\downarrow$ & SSIM $\uparrow$ & PSNR $\uparrow$ & MSE $\downarrow$ \\
\midrule
pixelNeRF~\cite{pixelnerf} & 0.591 & 0.460 & 13.91 & 0.0440 \\
IBRNet~\cite{wang2021ibrnet} & 0.532 & 0.484 & 15.99 & 0.0280 \\
GPNR~\cite{suhail2022generalizable} & \sbest 0.459 & \sbest 0.748 & \sbest 18.55 & \sbest 0.0165 \\
Ours & \best \textbf{0.262} & \best \textbf{0.839} & \best \textbf{21.38} & \best\textbf{0.0110}  \\
\bottomrule
\end{tabular}

}
\caption{
\textbf{Novel view rendering performance on RealEstate10K.} Our method outperforms all baselines on all metrics. 
}
\vspace{-0.05in}
\label{tab:realestate}
\end{table}

\begin{table}[t]
\centering

\scalebox{0.9}{
\begin{tabular}{lccccc}
\toprule
Method & LPIPS $\downarrow$ & SSIM $\uparrow$ & PSNR $\uparrow$ & MSE $\downarrow$ \\
\midrule
pixelNeRF~\cite{pixelnerf} &  0.628 & 0.464 & 16.48 & 0.0275\\
IBRNet~\cite{wang2021ibrnet} & \sbest0.385 & 0.513 & \sbest19.24 & \sbest0.0167 \\
GPNR~\cite{suhail2022generalizable} &  0.558 &  \sbest0.719 & 17.57 & 0.0218\\
Ours &  \best\textbf{0.364} & \best\textbf{0.781} & \best\textbf{23.63} & \best\textbf{0.0074} \\
\bottomrule
\end{tabular}
}
\vspace{-5pt}
\caption{
\textbf{Novel view rendering performance on ACID.} Our method outperforms all baselines on all metrics.
}
\vspace{-0.05in}
\label{tab:acid}
\vspace{-10pt}
\end{table}

\subsection{Outdoor Scene Neural Rendering}
\label{sect:outdoor}

We further evaluate on outdoor scenes with potentially unbounded depth.

\myparagraph{Qualitative Results.} We illustrate qualitative results in \fig{fig:acid}. In comparison to the baselines, our approach is able to more accurately reconstruct the geometry, and is able to synthesize multi-view consistent renderings from two large baseline views.

\myparagraph{Quantitative Results.} 
Similar to indoor scenes, our approach also outperforms all baselines in terms of all metrics on outdoor scenes, see \tbl{tab:acid}.

\subsection{Ablations and Analysis}
\label{sect:analysis}

We next analyze and ablate individual components of our approach.
We use the RealEstate10k dataset for these experiments.

\myparagraph{Ablations.}

We evaluate the importance of different components of our method in \tbl{tab:ablations}. 
The ``Base Model'' corresponds to a vanilla architecture that does not include some of our proposed contributions. 
It samples points uniformly in 3D, instead of our proposed 2D epipolar line sampling. 
It uses a monocular encoder instead of our proposed multi-view encoder, and does not use correspondence matching across views for refining the geometry. 
It also does not use the regularization loss for multi-view consistency or any data augmentation during  training. 
We find that all components of our approach are essential for high-quality performance. 
The results in \tbl{tab:ablations} show that sampling in  3D  sub-optimally uses the information in the feature maps, that our multi-view encoder and cross-image correspondence matching can compute features that better encode the 3D scene structure compared to monocular encoders, and that  data augmentation helps with generalization. 
While we found that the incorporation of the regularization loss led to a slight decrease in PSNR, we found that it improved multi-view consistency in the rendered video results, and also improved both LPIPS and SSIM perceptual metrics. 

\begin{table}[t]
\centering
\resizebox{\columnwidth}{!}{%

\footnotesize
\begin{tabular}{lcccc}
\toprule
\bf Models & LPIPS$\downarrow$ & SSIM$\uparrow$ &  PSNR$\uparrow$  & MSE$\downarrow$\\
\midrule
  Base Model & 0.452 & 0.735 & 18.11  & 0.0201 \\
  + 2D Sampling & 0.428 & 0.762 & 19.02 & 0.0159  \\
  + Cross Correspondence & 0.415 & 0.766  & 19.52  & 0.0142  \\
  + Multiview Encoder & 0.361 &  0.794 & 20.43  & 0.0132 \\
  + Regularization Loss & 0.358  & 0.808  & 19.84 & 0.0139 \\
  + Data Aug & 0.262  & 0.839  & 21.38  & 0.0110 \\
\bottomrule
\end{tabular}

}
\vspace{-5pt}
\caption{ \textbf{Ablations.}
All components of our proposed method are essential for high-quality novel view synthesis.
}
\vspace{-0.05in}
\label{tab:ablations}
\vspace{-10pt}
\end{table}

\begin{figure}[t]
    \centering
    \includegraphics[width=0.85\linewidth]{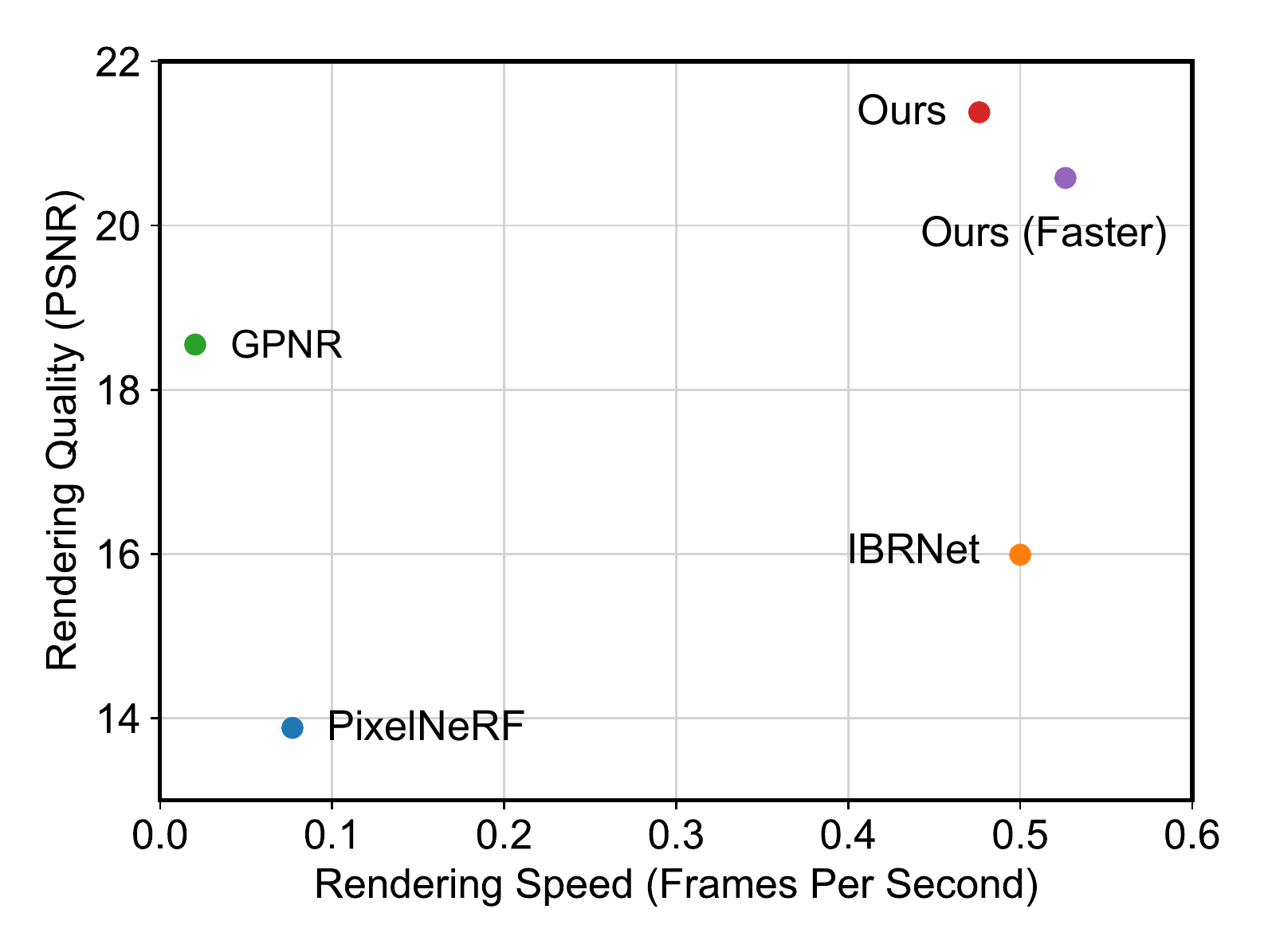}
    \vspace{-10pt}
    \caption{\textbf{FPS vs PSNR.} Our approach strikes the best trade-off between rendering quality and rendering speed. 
    We can further reduce the number of Epipolar samples (``Ours (Faster)''), which makes our method faster than all baselines, while still significantly outperforming them in terms of rendering quality.}
    \label{fig:time_hist}
    \vspace{-15pt}
\end{figure}

\myparagraph{Speed.} Next, in \fig{fig:time_hist}, we study the relationship between rendering quality and rendering speed for all approaches. 
Our lightweight approach achieves the best trade-off, significantly outperforming all methods in terms of rendering quality, 
while being at-par with the most efficient baseline. 
By reducing the number of sampled epipolar points from $64$ to $48$ samples per image, we can further speed up our approach, outperforming all baselines both in terms of rendering speed and image quality.

\begin{figure}[t]
    \centering
    \includegraphics[width=0.85\linewidth]{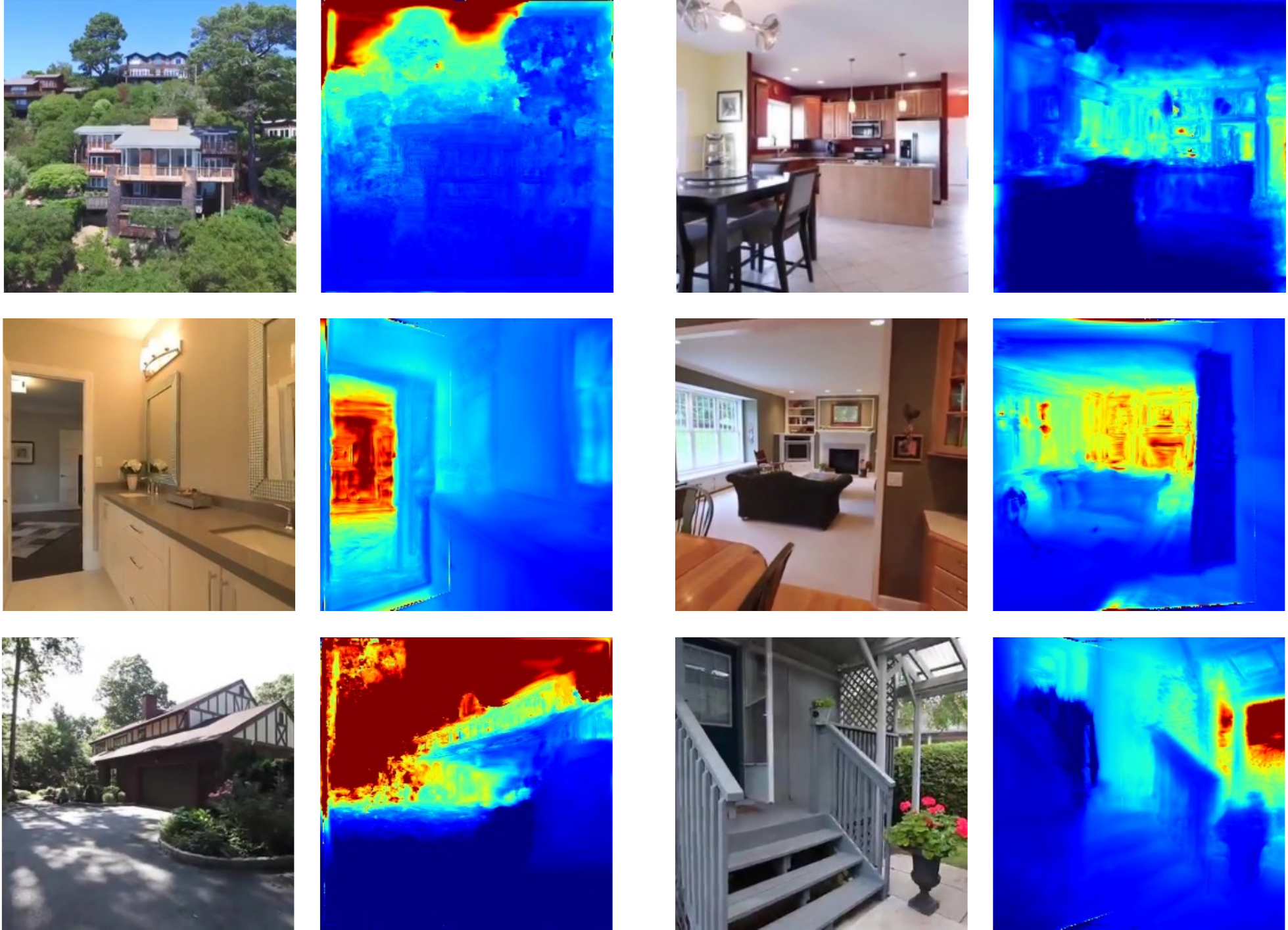}
    \caption{\textbf{Visualization of Epipolar Attention Weights.} The expected value of the depths of the epipolar samples under the attention weights can be seen as a depth proxy. 
    As our renderer is \emph{not} a volume renderer, these attention weights need not exactly correspond to the actual depth for correct renderings.}
    \label{fig:depth}
    \vspace{-15pt}
\end{figure}

\myparagraph{Epipolar Attention.} Finally, we visualize the underlying epipolar attention weights learned by our approach in \fig{fig:depth}. The expected value of the depths of the epipolar samples can be seen as a proxy depth and corresponds roughly to the underlying geometry of the scene.
This enables us to analyze the learned computation of our renderer.

\subsection{Novel View Synthesis from Unposed Images}
\label{sect:in_the_wild}
\begin{figure}[t]
    \centering
    \includegraphics[width=\linewidth]{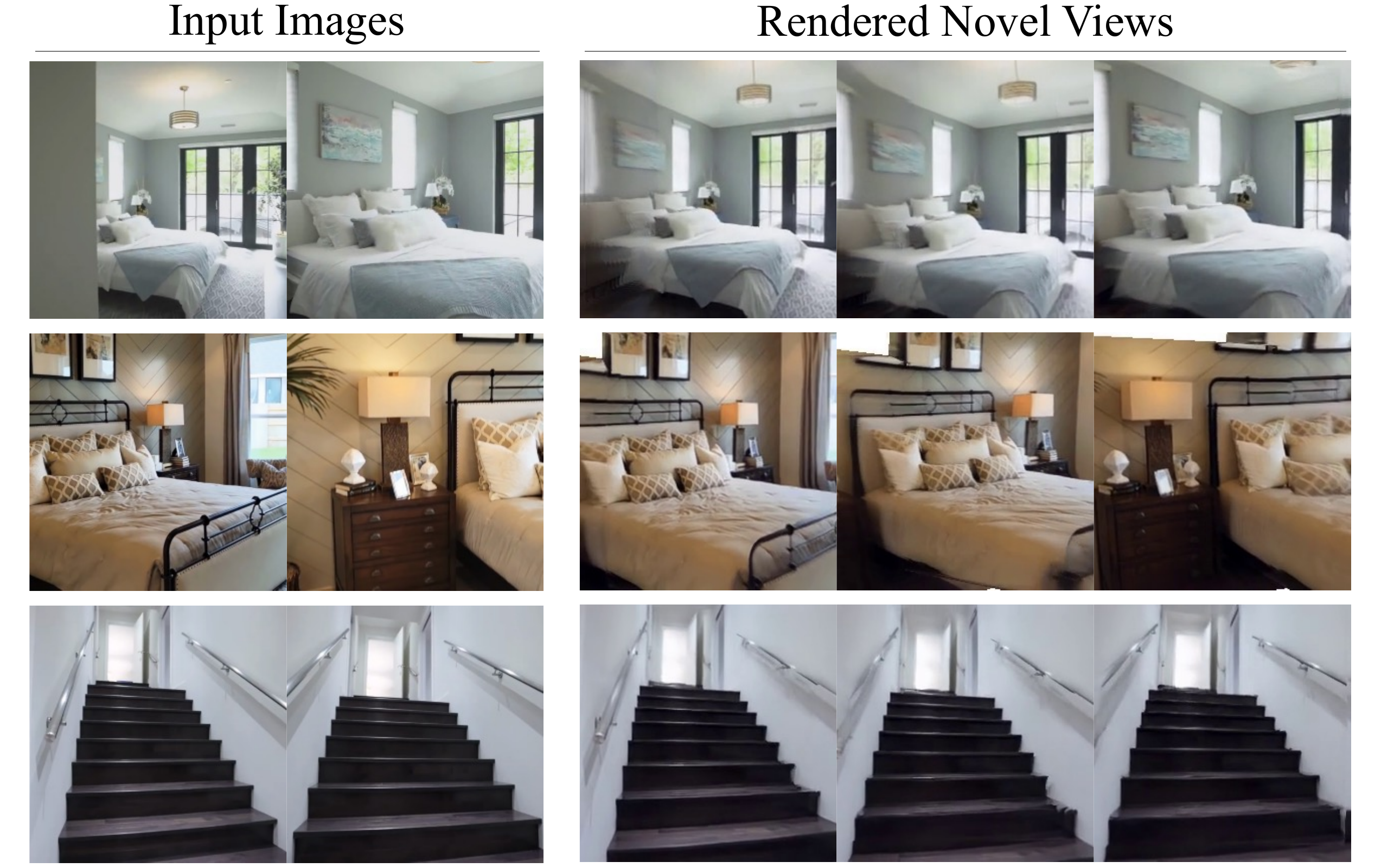}
    \caption{\textbf{Novel View Synthesis from Unposed Images.} Our approach can also render novel views using two unposed images captured in the wild.
    Note that parts of the scene only visible in one of the images can be correctly rendered from novel viewpoints.}
    \label{fig:wild}
    \vspace{-15pt}
\end{figure}

Our method uses a wide-baseline stereo image as input with known relative pose between them. 
We show that our method can perform novel view synthesis even without the knowledge of this relative pose information. 
In this case, we utilize SuperGlue~\cite{sarlin2020superglue} to  compute reliable pixel correspondences between the input images.
Since we do not know the camera intrinsics for in-the-wild images, we use the average intrinsics of the RealEstate10k dataset and compute the Essential matrix from the correspondences using RANSAC~\cite{fischler1981random}. 
We then compute the pose information from the essential matrix~\cite{horn1990recovering} and use it as input for our method. 
Note that the recovered translation is only defined up to a scale. 
\fig{fig:wild} demonstrates results on some in-the-wild scenes using images from the internet. 
Even in this unposed setting, our method can reason about the geometry of the scene by aggregating information across the sparse input views. 
This is an extremely challenging setting, and existing approaches for novel view synthesis from sparse views do not demonstrate any results on unposed images.

\section{Discussion}
While we have presented the first approach for novel view synthesis of scenes from very sparse input views, our approach still has several limitations. 
Our rendering results are not at the same quality as those obtained by methods that optimize on single scenes using more images.
Learning priors that enable novel view synthesis from sparse views is a significantly more challenging problem compared to using a large number of input images, where 3D points are regularly observed in many images. 
Our approach takes a step towards photorealistic renderings of scenes using only sparse views.
As our approach relies on learned priors, it does not generalize well to new scenes with very different appearances compared to the training scenes. 
However, our efficient approach lends itself to large-scale training on diverse datasets,  in turn enabling reconstruction of diverse scenes. 
Finally, while our method, in theory, can be extended to take more than two input views, we have only experimented with two views as a first step towards very sparse multi-view neural rendering.

\section{Conclusion}
\label{sec:conclusion}

We introduce a method for implicit 3D reconstruction and novel view synthesis from a single, wide-baseline stereo pair, trained using only self-supervision from posed color images.
By leveraging a multi-view encoder, an image-space epipolar line feature sampling scheme, and a cross-attention based renderer, our method surpasses the quality of prior art on datasets of challenging scenes. 
Our method further strikes a compelling trade-off between rendering speed and quality, rendering novel views significantly faster than most prior methods. 
Meanwhile, leveraging epipolar line geometry strikes a compelling trade-off between structured and generalist learning paradigms, enabling us to train our method on real-world datasets such as RealEstate10k.
We believe that this work will inspire the community towards further exploring the regime of extreme few-shot and generalizable novel view synthesis.

\vspace{3pt}
\noindent\textbf{Acknowledgements.}
This work was supported by the National Science Foundation under Grant No. 2211259, and by the Singapore DSTA under DST00OECI20300823 (New Representations for Vision). Yilun Du is supported by a NSF Graduate Research Fellowship.

{\small
\bibliographystyle{ieee_fullname}
\bibliography{references}
}

\ifarxiv \clearpage \appendix
\label{sec:appendix}

In this supplemental document, we provide experimental details of our method (\sect{sect:experiments}), additional comparisons with other baselines (\sect{sect:additional}), additional analysis of our approach (\sect{sect:appendix_results}), and derivation of epipolar correspondences in (\sect{sect:epipolar})\let\thefootnote\relax\footnotetext{$\dagger$ Equal Advising }.
Please refer to the project webpage for video results. 
\section{Experimental Details}
\label{sect:experiments}

We provide detailed experimental details necessary to reproduce the results listed in our paper.

\myparagraph{Dataset Details.} We use the download  script from \url{https://github.com/cashiwamochi/RealEstate10K_Downloader} to download 
videos in RealEstate and ACID datasets at $640 \times 480$ image resolution. 
Our datasets are smaller than the original ACID and RealEstate datasets because some of the listed YouTube URLs were not available anymore.

\myparagraph{Training Details.} 
We use a batch size of 48 and train our models using the Adam optimizer with a learning rate  of 5e-5.
We train on 4 Nvidia V100 GPUs for around $100$k iterations, which takes a total of 3 days. 
We do not use LPIPS and regularization losses for the first $30$k iterations. 
Both LPIPS and the regularization losses are computed across 32x32 patches of rendered images.  
Input frames are sampled so that are between 92 and 150 frames apart with intermediate frames rendered.

\myparagraph{Model Architecture.}
We utilize the VIT architecture from Ranftl~\etal\cite{ranftl2021vision}
 as our multi-view backbone. 
 We use the output feature maps of the last 2 RefineNet branches of the architecture as our features. 
The high-resolution feature map is obtained by applying a single convolutional layer with a kernel size of 3x3 with 64 channel dimensions.
We embed query tokens using a 2 layer MLP with hidden dimension of 128. We likewise obtain key vectors for cross-attention using a 2 layer MLP on input features. Attention values are computed using the dot product of key and query vectors, with dot product between vectors scaled by 1/16 for numerical stability.
For the second round of cross-attention, the output feature from the previous round of cross-attention is concatenated to each query token.
The MLP architecture used to decode RGB colors from pooled features is 3 layers in size with a hidden dimension of size 128.

\myparagraph{Evaluation Details} We use test scenes for evaluation in both RealEstate10k and ACID datasets. 
We use two frames 128 timesteps apart as the input to the methods and reconstruct an intermediate frame using the GT pose from the datasets.

\myparagraph{Neural Rendering of Unposed Images.} 
As mentioned in the main paper, we use SuperGlue~\cite{sarlin20superglue} to estimate correspondences between two unposed images, and then estimate the relative pose between them by computing the essential matrix. 
We use the average RealEstate10k intrinsic parameters. 
The recovered translation is only defined up to scale. 
We perform a grid search to find the best-performing scale offset.
We set the intrinsic matrix of unposed images to be the average focal length of scenes in RealEstate10k (225).

\begin{figure}[h]
    \centering
    \includegraphics[width=\linewidth]{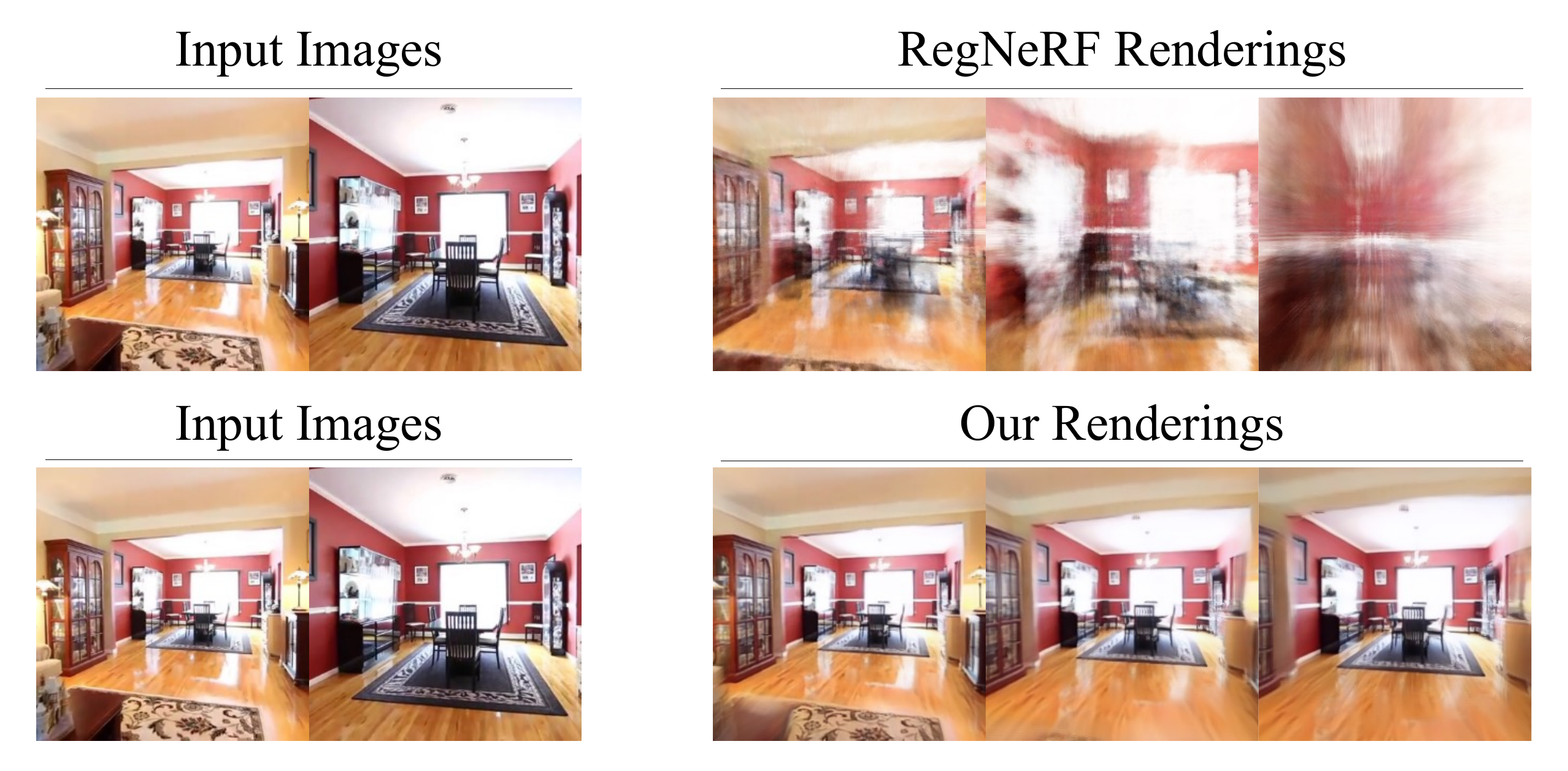}
    \caption{\textbf{Visualization of RegNeRF Renderings.} Comparison of RegNeRF renderings (top) with renderings of our method (bottom).}
    \label{fig:regnerf}
\end{figure}

\begin{table}[h]
\centering
\scalebox{0.75}{%

\begin{tabular}{lccccc}
\toprule
Method & LPIPS $\downarrow$ & SSIM $\uparrow$ & PSNR $\uparrow$ & MSE $\downarrow$ \\
\midrule
RegNeRF (Single Scene)~\cite{niemeyer2022regnerf} &  0.669 &  0.491 & 11.59  & 0.0741 \\
Ours (Single Scene)&  0.209 &  0.657 & 20.12  & 0.0102 \\
\midrule
pixelNeRF~\cite{pixelnerf} & 0.591 & 0.460 & 13.91 & 0.0440  \\
StereoNeRF~\cite{SRF} &  0.604  &  0.486 & 15.40 & 0.0318 \\
GeoNeRF~\cite{johari2022geonerf} & 0.541    & 0.511   &  16.65 & 0.0209  \\
IBRNet~\cite{wang2021ibrnet} & 0.532 & 0.484 & 15.99 & 0.0280 \\
GPNR~\cite{suhail2022generalizable} & \sbest 0.459 & \sbest 0.748 & \sbest 18.55 & \sbest 0.0165 \\
Ours & \best \textbf{0.262} & \best \textbf{0.839} & \best \textbf{21.38} & \best\textbf{0.0110}  \\
\midrule
\bottomrule
\end{tabular}
}
\caption{
\textbf{Extended table of Novel view rendering performance on RealEstate10K.} Our method outperforms all baselines on all metrics. RegNeRF results are reported for one evaluation scene (as the method requires a separate model to be fit per scene).
}
\vspace{-0.05in}
\label{tab:realestate_extended}
\end{table}

\section{Additional Baseline Comparisons}
\label{sect:additional}

We further compare with RegNeRF~\cite{niemeyer2022regnerf}
, StereoNeRF~\cite{SRF}, GeoNeRF~\cite{johari2022geonerf}. 
Quantitative comparisons with all baselines can be found in \tbl{tab:realestate_extended}. 
We significantly outperform these baselines. 
Since RegNeRF is scene-specific, we perform this evaluation on one test scene of the RealEstate10k dataset. 
RegNeRF takes several hours to compute the 3D reconstruction for a single scene, unlike our approach, where only a single forward pass is used. 
Qualitative comparisons can be found in \fig{fig:regnerf}.
Our method can better reconstruct the 3D scene structure as it learns a prior over scenes.

\section{Additional Analysis Results}
\label{sect:appendix_results}

We provide further analysis of our approach below.

\myparagraph{Performance with Epipolar Samples.} In \tbl{tbl:appendix_epipolar}, we illustrate the effects of using uniform samples on the epipolar lines compared to a large number of volumetric samples. A very large number of volumetric samples still does not match the underlying performance of epipolar samples.

\myparagraph{Multiview Encoder Ablations.} We qualitatively illustrate the ablation of adding a multiview compared to a single image encoder in \fig{fig:appendix_multiview_ablation}.

\myparagraph{Results on Varying Context Views} We illustrate how we can render our approach with a different number of views in \tbl{tbl:appendix_context}. We qualitatively illustrate rendering results with a different number of context views in \fig{fig:render_multiview}. Our renderer improves performance with a larger number of context views.

\myparagraph{Results on Varying Baseline Size.} In \tbl{tbl:appendix_baseline}, we illustrate rendering performance as we change the underlying baseline from which our approach is rendered. We find that as we decrease the baseline (distance) between frames, the underlying rendering performance improves.

\begin{figure*}[t]
    \centering
    \includegraphics[width=1\linewidth]{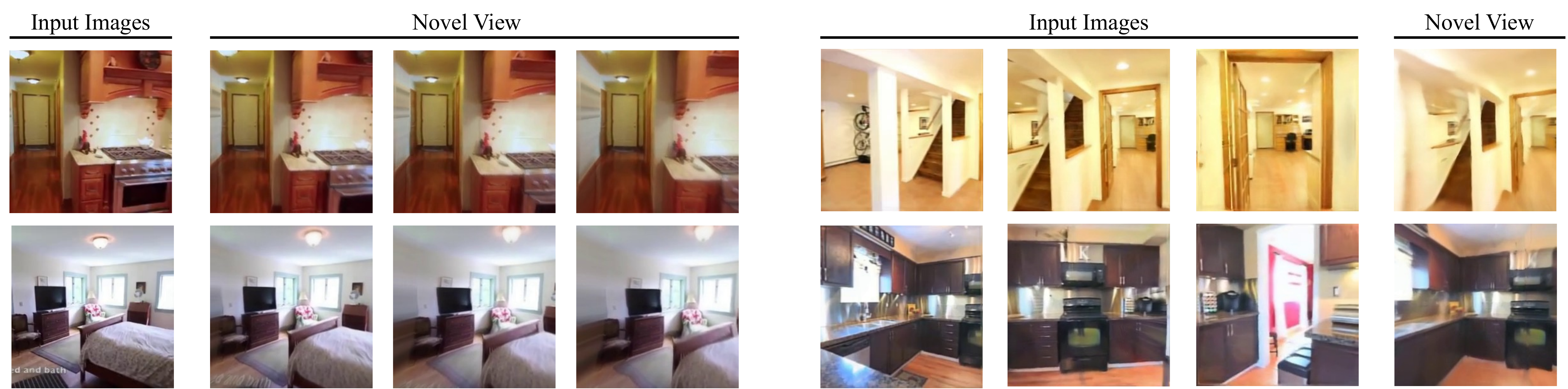}
    \caption{\textbf{Rendering with Different Context Views.} Visualization of rendering results when rendering with multiple context views.}
    \label{fig:render_multiview}
\end{figure*}

\begin{table}[t]
\centering
\begin{tabular}{l|ccc|c}
\toprule
3D Samples & 64 & 128 & 192 & Epi. Samples  \\
\midrule
  PSNR $\uparrow$ & 19.29  &  20.35  & 20.60 & \textbf{21.38}  \\
  SSIM $\uparrow$ &  0.769 &  0.778  & 0.790 & \textbf{0.839} \\
  LPIPS $\downarrow$ & 0.319  & 0.284 & 0.273  & \textbf{0.262}\\
\bottomrule
\end{tabular}
\vspace{-5pt}
\caption{\textbf{Rendering Results with Volumetric Samples.} Performance of rendering as a function of the number of volumetric samples used. A large number of volumetric samples still does not match epipolar samples.}
\vspace{-5pt}
\label{tbl:appendix_epipolar}
\end{table}

\begin{table}[t]
\centering
\begin{tabular}{lcccc}
\toprule
\bf Views & 1 & 2 & 3\\
\midrule
  PSNR $\uparrow$ & 18.48 & 21.38 & 22.29  \\
  SSIM $\uparrow$ &  0.700 & 0.839 & 0.848 \\
  LPIPS $\downarrow$ & 0.357 & 0.262 & 0.251 \\
\bottomrule
\end{tabular}
\vspace{-5pt}
\caption{\textbf{Rendering Results with Different Context Views.} Performance of rendering as a function of number of context views used.}
\vspace{-5pt}
\label{tbl:appendix_context}
\end{table}

\begin{table}[t]
\centering
\begin{tabular}{lcccc}
\toprule
\bf Baseline & 32 & 64 &  96 & 128\\
\midrule
  PSNR $\uparrow$ & 26.24  & 22.50 & 21.93 & 21.38  \\
  SSIM $\uparrow$ & 0.915 & 0.852 & 0.845 & 0.839 \\
  LPIPS $\downarrow$ & 0.149 & 0.223 & 0.246 & 0.262\\
\bottomrule
\end{tabular}
\vspace{-5pt}
\caption{\textbf{Rendering Results with Baseline Changes.} Performance of rendering as a function of change of baseline. Smaller baselines induce higher quality renderings.}
\vspace{-5pt}
\label{tbl:appendix_baseline}
\end{table}

\begin{figure}[t]
    \centering
    \includegraphics[width=1\linewidth]{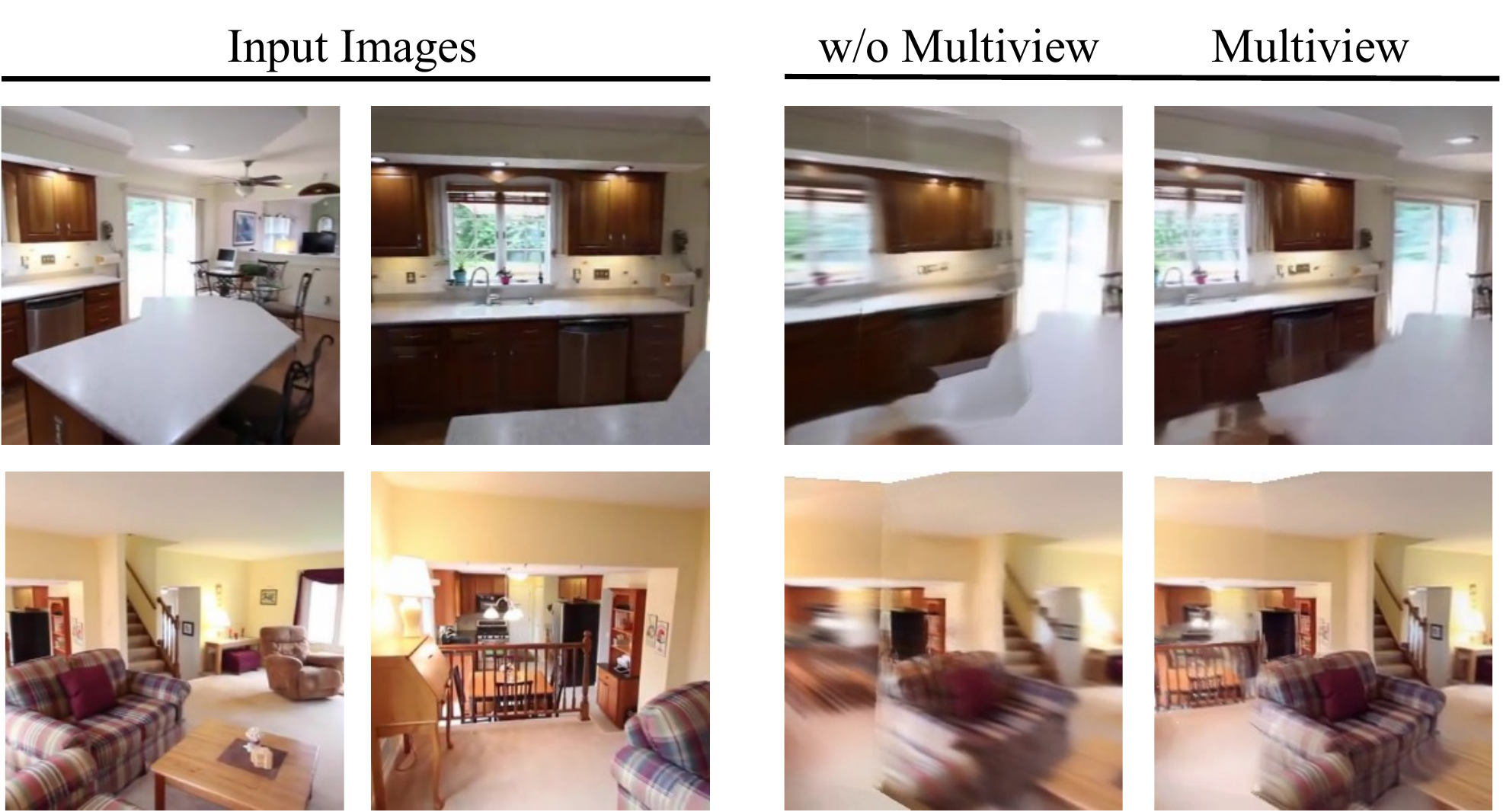}
    \caption{\textbf{Ablation of Multiview Encoder.} Qualitative visualization of rendering results when removing or adding a multiview encoder.}
    \label{fig:appendix_multiview_ablation}
\end{figure}

\section{Triangulation}
\label{sect:epipolar}

Here, we provide details on computing 3D points using triangulation.
For a pixel coordinate in the context image $(u', v')$, we may solve for its corresponding 3D point via:
\begin{align}
    \label{eq:3d_point_from_epi_point}
    l^* = \argmin_l \| \pi_t(\mathbf{o}_i  + l \cdot \mathbf{R}^{-1}_i \mathbf{K}^{-1}_i [u', v', 1]) - \mathbf{u}_t\|^2_2,
\end{align}
where $\mathbf{o}_i$ is the camera origin of the respective context image, $\pi_t(\cdot)$ denotes projection onto the target camera, and $\mathbf{u}_t$ is the pixel coordinate of the target ray we aim to render. 
The 3D point $\mathbf{p}^*$ can then be obtained as $\mathbf{p}^* = \mathbf{o}_i  + l^* \cdot \mathbf{R}^{-1}_i \mathbf{K}^{-1}_i [u', v', 1]$, and its depth in the context camera can be obtained as the $z$-coordinate of the point in the context camera's coordinates. 
Let $\mathbf{r}_i$ denotes the normalized ray direction $\mathbf{R}^{-1}_i \mathbf{K}^{-1}_i [u', v', 1]$. 
The closed form solution can be represented as:
\begin{align*}
    l^* &= \frac{u \cdot \mathbf{o}_i[z] - c_x \mathbf{o}_i[z] - f_x \mathbf{o}_i[x]}{f_x \mathbf{r}_i[x] + c_x \mathbf{r}_i[x] - u \mathbf{r}_i[z]} \\ 
    &= \frac{v \cdot \mathbf{o}_i[z] - c_y \mathbf{o}_i[z] - f_y \mathbf{o}_i[y]}{f_y \mathbf{r}_i[y] + c_y \mathbf{r}_i[y] - u \mathbf{r}_i[z]},
\end{align*}
where $\textbf{K}= \left[
\begin{smallmatrix}
f_x & 0 & c_x \\
 0 & f_y & c_y \\
 0 & 0 & 1 \\
\end{smallmatrix}\right]$
.

 \fi

\end{document}